\documentclass[twoside]{article}

\usepackage[accepted]{aistats2019}
\usepackage{booktabs} 
\usepackage{algorithmic}
\usepackage{amsmath}
\usepackage{float}
\usepackage{amssymb}
\usepackage{amsthm}
\usepackage{subcaption}
\usepackage{algorithm}
\usepackage{tabularx}
\newtheorem{theorem}{Theorem}[section]
\newtheorem{corollary}{Corollary}[theorem]

\newtheorem{definition}{Definition}[section]
\DeclareMathOperator*{\argmin}{arg\,min}
\newcommand\T{\rule{0pt}{2.6ex}}       
\usepackage{verbatim}
\usepackage[section]{placeins}
\usepackage{graphicx}
\usepackage{verbatim}
\usepackage{cite}

\makeatletter
\AtBeginDocument{%
  \expandafter\renewcommand\expandafter\subsection\expandafter{%
    \expandafter\@fb@secFB\subsection
  }%
}
\makeatother

\begin{document}

%

%

\twocolumn[

\aistatstitle{Revisiting hard thresholding for DNN pruning.}

\aistatsauthor{ Konstantinos Pitas \And Mike Davies  \And Pierre Vandergheynst  }

\aistatsaddress{ konstanstinos.pitas@epfl.ch \And Mike.Davies@ed.ac.uk \And pierre.vandergheynst@epfl.ch } 

]

\begin{abstract}
The most common method for DNN pruning is hard thresholding of network weights, followed by retraining to recover any lost accuracy. Recently developed smart pruning algorithms use the DNN response over the training set for a variety of cost functions to determine redundant network weights, leading to less accuracy degradation and possibly less retraining time. For experiments on the total pruning time (pruning time + retraining time) we show that hard thresholding followed by retraining remains the most efficient way of reducing the number of network parameters. However smart pruning algorithms still have advantages when retraining is not possible. In this context we propose a novel smart pruning algorithm based on difference of convex functions optimisation and show that it is often orders of magnitude faster than competing approaches while achieving the lowest classification accuracy degradation. Furthermore we investigate theoretically the effect of hard thresholding on DNN accuracy. We show that accuracy degradation increases with remaining network depth from the pruned layer. We also discover a link between the latent dimensionality of the training data manifold and network robustness to hard thresholding.  
\end{abstract}

\section{Introduction}
Deep neural networks have achieved state-of-the art results in a number of machine learning tasks \cite{lecun2015deep}. Training such networks is computationally intensive and often requires dedicated and expensive hardware. Furthermore, the resulting networks often require a considerable amount of memory to be stored. Using a Pascal Titan X GPU the popular AlexNet and VGG-16 models require 13 hours and 7 days, respectively, to train, while requiring 200MB and 600MB, respectively, to store. The large memory requirements limit the use of DNNs in embedded systems and portable devices such as smartphones, which are now ubiquitous. 

A number of approaches have been proposed to reduce the DNN size during training time, often with little or no degradation to classification performance. Approaches include introducing bayesian, sparsity-inducing priors \cite{louizos2017bayesian}, \cite{blundell2015weight}, \cite{molchanov2017variational}, \cite{dai2018compressing} and binarization \cite{hou2016loss} \cite{courbariaux2016binarized}.Other methods include the hashing trick used in \cite{chen2015compressing}, tensorisation \cite{novikov2015tensorizing} and efficient matrix factorisations \cite{yang2015deep}.

There has also been work in reducing the number of network parameters after training. Some first work in this direction has been \cite{kim2015compression} \cite{han2015deep} \cite{han2015learning} where hard thresholding is applied to network weights. However weight magnitude is a mediocre predictor for determining whether a weight is essential to network accuracy and following hard thresholding network accuracy often falls significantly. Therefore hard thresholding often has to be followed by a retraining procedure. Ideally one would want to remove parameters by minimizing the reduction in network accuracy over the training set, but this is computationally prohibitive. Smart pruning algorithms aim to overcome this problem by minimizing a surrogate cost function while pruning. In \cite{aghasi2016net}\cite{aghasi2018fast} the authors propose a convexified layerwise pruning algorithm termed Net-Trim, that minimizes the $l_2$ norm between pruned and unpruned layer output representations. The authors in \cite{dong2017learning} propose LOBS, an algorithm for layerwise pruning by approximating the loss function using a second order Taylor expansion. In \cite{baykal2018data} the authors use a coreset approach to prune individual layers and enforce an $l_{2}$ norm penalty between pruned and unpruned layer output representations with high probability.

Given that DNNs represent a highly non-linear function it is also of independent interest to explore theoretically what is the effect of pruning on DNN accuracy. Hard thresholding a neural network layer introduces a perturbation to the latent signal representations generated by that layer. As the pertubated signal passes through layers of non-linear projections, the perturbation could become arbitrarily large. DNN robustness to hidden layer perturbations has been investigated for random noise in \cite{raghu2016expressive} and for the case of pruning in \cite{aghasi2016net} and \cite{dong2017learning}. In the last two works the authors conduct a theoretical analysis using the Lipschitz properties of DNNs showing the stability of the latent representations, over the training set, after pruning by assuming that $||\boldsymbol{W}||_2=1$ where $\boldsymbol{W}$ are the weights at a given layer. The Lipschitz properties of DNNs have also been used to analyze their generalization error (GE), \cite{sokolic2017robust} \cite{bartlett2017spectrally} \cite{neyshabur2017pac}. Very close to our work \cite{bartlett2017spectrally} show that the difficulty of a training set can be captured by normalizing the classification margins of training samples with a "spectral complexity" term that includes the spectral norms of the layer weights. Randomly labelled data as opposed to real labelled data produce networks with much smaller normalized margins, reflecting the difficulty of the randomly labelled dataset. 

For the practical pruning section including with and without retraining we focus on pruning only the fully connected non-linear layers of a network. Two of the proposed algorithms are only applicable in fully connected layers. Furthermore fully connected layers often include $90\%$ of network parameters, and are the most redundant of layers in most architectures. Finally the softmax linear layer can be readily compressed using an SVD decomposition \cite{yang2015deep}.

\subsection{Contributions}
In this work we make the following contributions 

\begin{itemize}
\item We perform pruning and retraining experiments for various datasets and architectures and discover that hard thresholding is the most efficient pruning pipeline often taking just a fraction of the total pruning and retraining time of other approaches. For the case without retraining smart pruning approaches lead to significant accuracy gains for the same level of sparsity, especially for DNNs with only fully connected layers.
\item We propose a novel smart pruning algorithm called FeTa that can be cast as a difference of convex functions problem, and has an efficient solution. For a fully connected layer with input dimension $d_1$, output dimension $d_2$ and $N$ training samples, the time complexity of our iterative algorithm scales like $\mathcal{O}(K(N+\frac{Nk}{N+\sqrt{k}}) \text{log}(\frac{1}{\epsilon}) d_1 d_2)$, where $\epsilon$ is the precision of the solution, $k$ is related to the Lipschitz and strong convexity constants, $d_2 \ll d_1$ and $K$ is the outer iteration number. Competing approaches scale like $\mathcal{O}(d_1^2)$ and $\mathcal{O}(d_1^3)$ and we conduct experiments showing that our algorithm leads to higher accuracy in the pruned DNN and is often orders of magnitude faster.
\item We build upon the work of \cite{sokolic2017robust} to bound the GE of a DNN for the case of hard thresholding of hidden layer weights. In contrast to the analysis of \cite{aghasi2016net} and \cite{dong2017learning} our analysis correctly predicts that for the same level of sparsity, accuracy degrades with the remaining depth of the thresholded layer. 
\item Our theoretical analysis also predicts that a DNN trained on a data manifold with lower intrinsic dimensionality will be more robust to hard thresholding. We provide empirical evidense that validates this prediction. 
\end{itemize}
. 

\subsection{Notation and Definitions}
We use the following notation in the sequel: matrices, column vectors, scalars and sets are denoted by boldface upper-case letters ($\boldsymbol{X}$), boldface lower-case letters ($\boldsymbol{x}$), italic letters ($x$) and calligraphic upper-case letters ($\mathcal{X}$), respectively. The covering number of $\mathcal{X}$ with $d$-metric balls of radius $\rho$ is denoted by $\mathcal{N}(\mathcal{X};d,\rho)$. A $C_M$-regular $k$-dimensional manifold, where $C_M$ is a constant that captures "intrinsic" properties, is one that has a covering number $\mathcal{N}(\mathcal{X};d,\rho)=(\frac{C_M}{\rho})^k$. We define $\text{nnz}(\cdot)$ as the number of non-zeros of a vector or matrix, $\rho(x)=\textbf{\text{max}}(0,x)$ as the rectifier non-linearity, $I(\cdot)$ as the elementwise indicator function, and $\odot$ as the Hadamard product.

\section{Retraining Experiments}

We consider a classification problem, where we observe a vector $\boldsymbol{x} \in \mathcal{X} \subseteq \mathbb{R}^N$ that has a corresponding class label $y \in \mathcal{Y}$. The set $\mathcal{X}$ is called the input space, $\mathcal{Y} = \{1,2,...,N_{\mathcal{Y}}\}$ is called the label space and $N_{\mathcal{Y}}$ denotes the number of classes. The samples space is denoted by $\mathcal{S}=\mathcal{X} \times \mathcal{Y}$ and an element of $\mathcal{S}$ is denoted by $s = (\boldsymbol{x},y)$. We assume that samples from $\mathcal{S}$ are drawn according to a probability distribution $P$ defined on $\mathcal{S}$. A training set of $m$ samples drawn from $P$ is denoted by $S_m = \{s_i\}^m_{i=1}=\{(\boldsymbol{x}_i,y_i)\}^m_{i=1}$.

We consider DNN classifiers defined as

\begin{equation}
g(\boldsymbol{x}) = \max_{i \in [N_y] } (f(\boldsymbol{x}))_i ,
\end{equation}

where $(f(\boldsymbol{x}))_i$ is the $i-$th element of $N_{y}$ dimensional output of a DNN $f:\mathbb{R}^N \rightarrow \mathbb{R}^{N_y}$. We assume that $f(\boldsymbol{x})$ is composed of $L$ layers

\begin{equation}
f(\boldsymbol{x})=f_L(f_{L-1}(...f_1(\boldsymbol{x},\boldsymbol{W}_1),...\boldsymbol{W}_{L-1}),\boldsymbol{W}_L) ,
\end{equation}

where $f_l(\cdot,\boldsymbol{W}_l)$ represents the $l-$th layer with parameters $\boldsymbol{W}_l$, $l = 1,...,L$. The output of the $l-$th layer is denoted $\boldsymbol{z}^l$, i.e. $\boldsymbol{z}^l=f_l(\boldsymbol{z}^{l-1},\boldsymbol{W}_l)$. The input layer corresponds to $\boldsymbol{z}^{0} = \boldsymbol{x}$ and the output of the last layer is denoted by $\boldsymbol{z} = f(\boldsymbol{x})$.

For each training signal $\boldsymbol{x}_i \in \mathbb{R}^{N}$ we assume also that we have access to the inputs $\boldsymbol{a}_i \in \mathbb{R}^{d_1} $ and the outputs $\boldsymbol{b}_i \in \mathbb{R}^{d_2} $ of a fully connected layer to be pruned. For an unpruned weight matrix $\boldsymbol{W}_l$ the layer is defined by the equation $f_l(x,\boldsymbol{W}_l)=\rho(\boldsymbol{W}_l^Tx)$. We denote $\boldsymbol{A} = [\boldsymbol{a}_1,...,\boldsymbol{a}_m]$ the matrix concatenating all the latent input vectors $\boldsymbol{a}_j$ and $\boldsymbol{B} = [\boldsymbol{b}_1,...,\boldsymbol{b}_m]$ the matrix concatenating all the latent output vectors $\boldsymbol{b}_j$. We are looking for $\boldsymbol{U}$ a new pruned weight matrix such that the new layer will be defined as $f_l(x,\boldsymbol{U}_l)=\rho(\boldsymbol{U}_l^Tx)$.

\begin{table*}[t]
\caption{Test accuracy and total pruning time (hours (H), minutes (M), seconds (S)) for Mnist, Fashion Mnist and Cifar-10.} \label{tab:title2} 
\label{sample-table}
\vskip 0.15in
\begin{center}
\begin{small}
\begin{sc}
\begin{tabular}{lcccccccccccc} 
  \toprule
  {}     & \multicolumn{2}{c}{Dense Mnist} & \multicolumn{2}{c}{Conv Mnist} & \multicolumn{2}{c}{Dense F-Mnist} & \multicolumn{2}{c}{Conv F-Mnist} & \multicolumn{2}{c}{Dense Cifar} & \multicolumn{2}{c}{Conv Cifar} \\ 
  \midrule
  {}   & Time & Acc & Time & Acc & Time & Acc & Time & Acc & Time & Acc & Time & Acc \\ 
  \midrule
  Net-Trim & 3\text{m} & 94.5\% & 16\text{m} & 97\% & 15\text{m} & 87\% & 24\text{m} & 92\% & 13\text{m} & 45\% & 39\text{m} & 78\% \T\\   
  LOBS     & 3\text{m} & 96\% & 3\text{m} & 97\% & 3\text{m} & 87\% & 3\text{h} & 92\% & 1\text{h} & 45\% & 3\text{h} & 78\% \\
  Corenet  & 1\text{m} & 93\% & 3\text{m} & 97\% & 42\text{m} & 87\% & 18\text{m} & 92\% & 1\text{h} & 45\% & 53\text{m} & 76\% \\
  \midrule
  $\textbf{Threshold}$ & $\boldsymbol{26\text{s}}$ & 96\% & $\boldsymbol{0.5\text{s}}$ & 97\% & $\boldsymbol{4\text{s}}$ & 87\% & $\boldsymbol{22\text{s}}$ & 92\% & $\boldsymbol{3\text{s}}$ & 45\% & $\boldsymbol{44\text{s}}$ & 78\% \\
  \bottomrule
\end{tabular}
\end{sc}
\end{small}
\end{center}
\vskip -0.1in
\end{table*}

We compare three state of the art smart pruning techniques with hard thresholding for pruning feedforward neural networks. A smart pruning algorithm is defined as $\operatorname{SmartPrune}(\boldsymbol{W},\boldsymbol{A},\boldsymbol{B}) \rightarrow \boldsymbol{U}$ which takes as inputs $\boldsymbol{W},\boldsymbol{A},\boldsymbol{B}$ the original layer weights, the latent representation inputs and outputs to the layer and outputs $\boldsymbol{U}$ the new pruned weight matrix. We compare the following algorithms: (i) Hard thresholding defined as $\operatorname{HardThreshold}(\boldsymbol{W})=\boldsymbol{W} \odot I(|\boldsymbol{W}|>t)$, where $t$ is a positive constant. (ii) Net-Trim \cite{aghasi2018fast} (iii) LOBS \cite{dong2017learning} (iv) Corenet \cite{baykal2018data}. We refer to the original papers for details. We used implementations by the authors for Net-Trim and LOBS and created our own implementation of Corenet.  

Experiments were performed on three commonly used datasets, \textit{Mnist}\cite{lecun1998gradient}, \textit{FashionMnist}\cite{xiao2017fashion} and \textit{Cifar-10}\cite{krizhevsky2009learning}. For each dataset we used half the test set as a validation set. We also tested two different architectures for each dataset one fully connected and one convolutional. We plot the architectures tested in Figure 1. We refer to implementation details in APPENDIX C. 

For fully connected architectures both pruning and retraining was done on a MacBook Pro with CPU Intel Core i7 @ 2.8GHz and RAM 16GB 1600 MHz DDR3 using only the CPU. For convolutional architectures pruning was done on 48 Intel Xeon CPUs E5-2650 v4 @ 2.20GHz and retraining was done using a single GeForce GTX 1080 GPU. 

\begin{figure}[t!]
    \centering
    \begin{subfigure}[b]{0.12\linewidth}
        \includegraphics[scale=0.35]{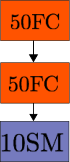}
        \caption{}
    \end{subfigure}
    \begin{subfigure}[b]{0.12\linewidth}
        \includegraphics[scale=0.35]{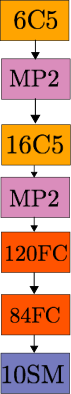}
        \caption{}
    \end{subfigure}
    \begin{subfigure}[b]{0.12\linewidth}
        \includegraphics[scale=0.35]{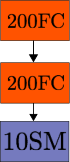}
        \caption{}
    \end{subfigure}
    \begin{subfigure}[b]{0.12\linewidth}
        \includegraphics[scale=0.35]{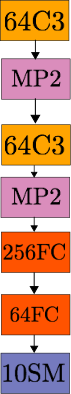}
        \caption{}
    \end{subfigure}
    \begin{subfigure}[b]{0.12\linewidth}
        \includegraphics[scale=0.35]{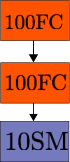}
        \caption{}
    \end{subfigure}
    \begin{subfigure}[b]{0.12\linewidth}
        \includegraphics[scale=0.35]{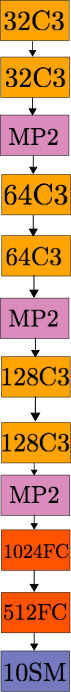}
        \caption{}
    \end{subfigure}
    \caption{\textbf{The DNN architectures that are tested}: a) Dense Mnist b) Convolutional Mnist c) Dense Fashion Mnist d) Convolutional Fashion Mnist e) Dense Cifar-10 f) Convolutional Cifar-10. We denote convolutional layers with $k$ ouput channels and $q \times q$ filter support by $k\text{C}q$. We denote fully connected layers with $k$ output dimensions by $k\text{FC}$. We denote max-pooling layers with $k \times k$ pooling regions by $\text{MP}k$ and softmax layers with $k$ output dimensions by $k\text{SM}$. }
\end{figure}

We first prune the fully connected nonlinear layers of the DNN architectures using the different algorithms to $90\%$ sparsity. We then retrain the entire pruned architecture until the lost accuracy is recovered. We present the results comparing total pruning time (pruning time + retraining time) for all architectures in Table 1. We see that, using retraining, hard thresholding is able to achieve the same accuracy in the pruned model compared to other approaches at only a fraction of the time. We see furthermore that Net-Trim is faster than Corenet and LOBS in the larger scale Cifar experiments. Finally LOBS is by far the most time consuming algorithm.  

We conclude that for the architectures tested and the computational resources that where allocated for pruning and retraining, smart pruning techniques represent a significant increase in time complexity without providing any benefit in the final network accuracy. We must note that while we consider the experimental setting which we consider the most realistic it might still be possible to see a benefit from smart pruning techniques if more efficient computational resources are allocated to smart pruning or less efficient resources are available for retraining. For example retraining a convolutional network using only CPUs might be less efficient than smart pruning. 

We determine that pruning accuracy vs pruning speed is a crucial parameter to be optimised by any smart pruning algorithm. In this context we derive in the next section a novel pruning algorithm that is often orders of magnitude faster than competing approaches while still achieving the highest network accuracy after pruning.

\section{The FeTa algorithm}
In this section we derive our novel smart pruning algorithm.

\subsection{Net-Trim}
We first make a detailed description of the Net-Trim algorithm \cite{aghasi2018fast} which will be the basis for our approach. For $\boldsymbol{U} \in \mathbb{R}^{d_1 \times d_2}, \boldsymbol{A} \in \mathbb{R}^{d_1 \times m},\boldsymbol{B} \in \mathbb{R}^{d_2 \times m},\epsilon \in \mathbb{R}^+$, Net-Trim aims to minimize

\begin{equation}
\min_{\boldsymbol{U}} ||\boldsymbol{U}||_1 \;\; \text{subject to} \;\; ||\rho(\boldsymbol{U}^{T}\boldsymbol{A})-\boldsymbol{B}||^2_F \leq \epsilon,
\end{equation}

where $\epsilon$ is a tolerance parameter. Intuitively Net-Trim tries to find a sparse matrix $\boldsymbol{U}$ such that the output of the pruned nonlinear layer $\rho(\boldsymbol{U}^{T}\boldsymbol{A})$ stays close in terms of the Frobenius norm to the original unpruned output $\boldsymbol{B}$. The optimisation above is non-convex and the authors of Net-Trim introduce a convexified formulation. For $\boldsymbol{U} \in \mathbb{R}^{d_1 \times d_2}, \boldsymbol{A} \in \mathbb{R}^{d_1 \times m},\boldsymbol{B} \in \mathbb{R}^{d_2 \times m},\epsilon \in \mathbb{R}^+$, and $\Omega \subseteq \{1,...,d_2\} \times \{1,...,m\}$, Net-Trim minimizes

\begin{equation}
\min_{\boldsymbol{U}} ||\boldsymbol{U}||_1 \;\; \text{subject to} \;\; 
\begin{cases}
& ||(\boldsymbol{U}^T\boldsymbol{A}-\boldsymbol{B})_{\Omega}||_F \leq \epsilon\\
& (\boldsymbol{U}^T\boldsymbol{A})_{\Omega^c} \leq \boldsymbol{V}_{\Omega^c},\\ 
\end{cases}
\end{equation}  

which is solved using the ADMM approach. There are two main issues with this formulation. One is time complexity as the ADMM implementation has to solve a Cholesky factorisation which scales like $\mathcal{O}(d_1^3)$ where $d_1$ is the layer input dimension. The second is space complexity as the algorithm has to create inequality constraints, and thus allocate memory, proportional to the number of data samples $\mathcal{O}(m)$. In \cite{aghasi2018fast} the space complexity under some assumptions has been shown to be of the order $\mathcal{O}(\text{nnz}(\boldsymbol{U})\log(\frac{m}{\text{nnz}(\boldsymbol{U}}) ))$. We note however that for large layers $\text{nnz}(\boldsymbol{U})$ can be higher than $m$ even for high sparsity levels.

\subsection{DC decomposition}
We aim to overcome the shortcomings of Net-Trim by reformulating the optimisation problem as a difference of convex functions problem. For $\boldsymbol{U} \in \mathbb{R}^{d_1 \times d_2}, \boldsymbol{a_j} \in \mathbb{R}^{d_1},\boldsymbol{b_j} \in \mathbb{R}^{d_2},\lambda \in \mathbb{R}^+$ we reformulate the optimisation problem that we want to solve as

\begin{equation}
\min_{\boldsymbol{U}} \frac{1}{m} \sum_{s_j \in \mathcal{S}_m}||\rho(\boldsymbol{U}^{T}\boldsymbol{a}_j)-\boldsymbol{b}_j||^2_2+ \lambda \Omega (\boldsymbol{U}),
\end{equation} 

where $\lambda$ is the sparsity parameter. The term $||\rho(\boldsymbol{U}^{T}\boldsymbol{a}_j)-\boldsymbol{b}_j||^2_2$ ensures that the nonlinear projection remains the same for training signals. The term $ \lambda \Omega (\boldsymbol{U}) $ is any convex regulariser which imposes the desired structure on the weight matrix $\boldsymbol{U}$. 

The objective in Equation 5 is non-convex. We show that the optimisation of this objective can be cast as a difference of convex functions (DC) problem. We assume just one training sample $\boldsymbol{x} \in \mathbb{R}^{N}$, for simplicity, with latent representations $\boldsymbol{a} \in \mathbb{R}^{d_1} $ and $\boldsymbol{b} \in \mathbb{R}^{d_2} $

\begin{equation}
\begin{split}
& ||\rho(\boldsymbol{U}^{T}\boldsymbol{a})-\boldsymbol{b}||^2_2+ \lambda\Omega (\boldsymbol{U}) \\
& = \sum_{i}[\rho(\boldsymbol{u_i}^{T}\boldsymbol{a})-\boldsymbol{b_i} ]^2+\lambda \Omega (\boldsymbol{U}) \\
& = \sum_{i}[\rho^{2}(\boldsymbol{u_i}^{T}\boldsymbol{a})-2\rho(\boldsymbol{u_i}^{T}\boldsymbol{a})\boldsymbol{b_i}+\boldsymbol{b_i}^{2} ]+\lambda \Omega (\boldsymbol{U}) \\
& = \sum_{i}[ \rho^{2}(\boldsymbol{u_i}^{T}\boldsymbol{a})+\boldsymbol{b_i}^{2} ]+\lambda \Omega (\boldsymbol{U}) +\sum_{i}[-2\boldsymbol{b_i}\rho(\boldsymbol{u_i}^{T}\boldsymbol{a})] \\
& = \sum_{i}[ \rho^{2}(\boldsymbol{u_i}^{T}\boldsymbol{a})+\boldsymbol{b_i}^{2} ]+\lambda \Omega (\boldsymbol{U}) \\
& +\sum_{\substack {i \\ b_i<0} }[-2\boldsymbol{b_i}\rho(\boldsymbol{u_i}^{T}\boldsymbol{a})] + \sum_{\substack {i \\ b_i\geq0} }[-2\boldsymbol{b_i}\rho(\boldsymbol{u_i}^{T}\boldsymbol{a})].\\
\end{split}
\end{equation}

Notice that after the split the first term ($b_i < 0$) is convex while the second ($b_i \geq 0$) is concave. We note that $b_i \geq 0$ by definition of the ReLu and set

\begin{equation}
g(\boldsymbol{U};\boldsymbol{x}) = \sum_{i}[ \rho^{2}(\boldsymbol{u_i}^{T}\boldsymbol{a})+\boldsymbol{b_i}^{2} ],
\end{equation}

\begin{equation}
h(\boldsymbol{U};\boldsymbol{x}) = \sum_{\substack {i \\ b_i>0} }[2\boldsymbol{b_i}\rho(\boldsymbol{u_i}^{T}\boldsymbol{a})].
\end{equation}

Then by summing over all the samples we get

\begin{equation}
\begin{split}
f(\boldsymbol{U}) &= \sum_{j}g(\boldsymbol{U};\boldsymbol{x}_j)+\lambda \Omega (\boldsymbol{U}) - \sum_{j} h(\boldsymbol{U};\boldsymbol{x}_j) \\
          &= g(\boldsymbol{U})+\lambda \Omega (\boldsymbol{U}) - h(\boldsymbol{U}), \\
\end{split}
\end{equation}

which is difference of convex functions. The rectifier nonlinearity is non-smooth, but we can alleviate that by assuming a smooth approximation. A common choice for this task is $\rho(x) = \frac{1}{\theta}\text{log}(1+\text{exp}(\theta x))$, with $\theta$ a positive constant.

\subsection{Optimisation}
It is well known that DC programs have efficient optimisation algorithms. We propose to use the DCA algorithm \cite{tao1997convex}. DCA is an iterative algorithm that consists in solving, at each iteration, the convex optimisation problem obtained by linearizing $h(\cdot)$ (the non-convex part of $f = g - h$) around the current solution. Although DCA is only guaranteed to reach local minima the authors of \cite{tao1997convex} state that DCA often converges to the global minimum, and has been used succefully to optimise a fully connected DNN layer \cite{fawzi2015dictionary}. At iteration $k$ of DCA, the linearized optimisation problem is given by

\begin{equation}
\argmin_{\boldsymbol{U}}\{g(\boldsymbol{U})+\lambda \Omega (\boldsymbol{U})-Tr(\boldsymbol{U}^{T}\nabla h(\boldsymbol{U}^k))\},
\end{equation}

where $\boldsymbol{U}^{k}$ is the solution estimate at iteration $k$. The detailed procedure is then given in algorithms 1 and 2. We assume that the regulariser is convex but possibly non-smooth in which case the optimisation can be performed using proximal methods.

\begin{algorithm}[h!] 
\caption{FeTa (Fast and Efficient Trimming Algorithm)}
\label{alg:algorithm1}
\begin{algorithmic}[1]
\STATE Choose initial point: $\boldsymbol{U}^0$
\FOR {k = 1,...,K}
  \STATE Compute $C \gets \nabla h(\boldsymbol{U}^k)$.
  \STATE Use Algorithm 2 to solve the convex optimisation problem:
  \begin{equation}
  \boldsymbol{U}^{k+1} \gets \argmin_{\boldsymbol{U}}\{g(\boldsymbol{U})+\lambda \Omega (\boldsymbol{U})-Tr(\boldsymbol{U}^{T}C)\}
  \end{equation}
\ENDFOR
\STATE If $\boldsymbol{U}^{k+1} \approx \boldsymbol{U}^{k}$ return $\boldsymbol{U}^{k+1}$.
\end{algorithmic}
\end{algorithm}

\begin{algorithm}[h!] 
\caption{Acc-Prox-SVRG}
\label{alg:algorithm3}
\begin{algorithmic}[1]
\STATE \textbf{Initialization}: $\tilde{\boldsymbol{x} }_0 \gets \boldsymbol{U}^k , \beta , \eta $
\FOR {s = 1,...,S}
  \STATE $\tilde{\boldsymbol{u} } = \nabla g(\tilde{\boldsymbol{x} }_s)$
  \STATE $\boldsymbol{x}_1 = \boldsymbol{y}_1 = \tilde{\boldsymbol{x} }_s$
  \FOR {t = 1,2,...,T}
    \STATE Choose $(\boldsymbol{A}',\boldsymbol{B}')$ randomly chosen minibatch.
    \STATE $\boldsymbol{u}_t = \nabla g_{\boldsymbol{A}',\boldsymbol{B}'}(\boldsymbol{y}_t) - \nabla g_{\boldsymbol{A}',\boldsymbol{B}'}(\tilde{\boldsymbol{x} }_s)+\tilde{\boldsymbol{u} }$
    \STATE $\boldsymbol{x}_{t+1} = \text{prox}_{\eta h}(\boldsymbol{y}_t - \eta \boldsymbol{u}_t)$
    \STATE $\boldsymbol{y}_{t+1} = \boldsymbol{x}_{t+1} + \beta(\boldsymbol{x}_{t+1}-\boldsymbol{x}_t)$
  \ENDFOR
  \STATE $\tilde{\boldsymbol{x} }_{s+1} = \boldsymbol{x}_{T+1}$
\ENDFOR
\STATE Return $\boldsymbol{U}^{k+1} \gets \tilde{\boldsymbol{x} }_{S+1}$
\end{algorithmic}
\end{algorithm}

\begin{figure*}[t!]
    \centering
    \begin{subfigure}[b]{0.3\linewidth}
        \includegraphics[scale=0.35]{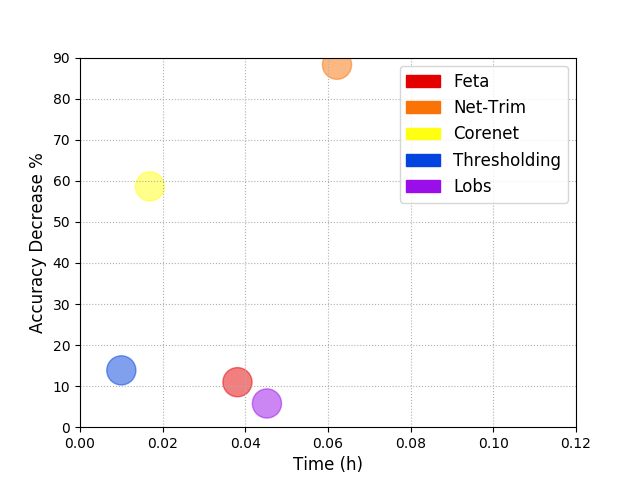}
        \caption{Dense Mnist}
    \end{subfigure}
    \begin{subfigure}[b]{0.3\linewidth}
        \includegraphics[scale=0.35]{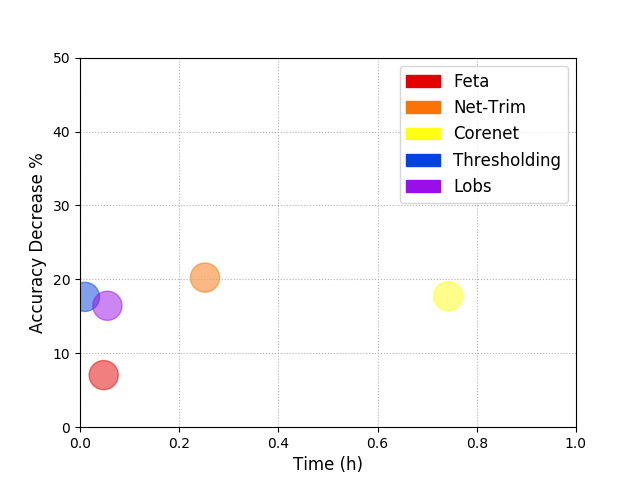}
        \caption{Dense Fashion Mnist}
    \end{subfigure}
    \begin{subfigure}[b]{0.3\linewidth}
        \includegraphics[scale=0.35]{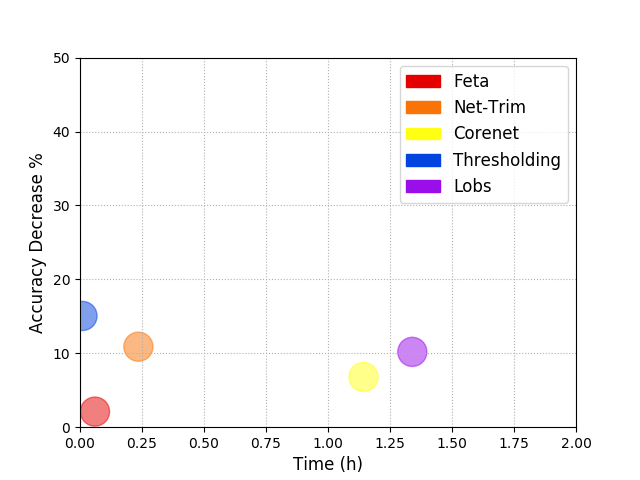}
        \caption{Dense Cifar}
    \end{subfigure}
    \begin{subfigure}[b]{0.3\linewidth}
        \includegraphics[scale=0.35]{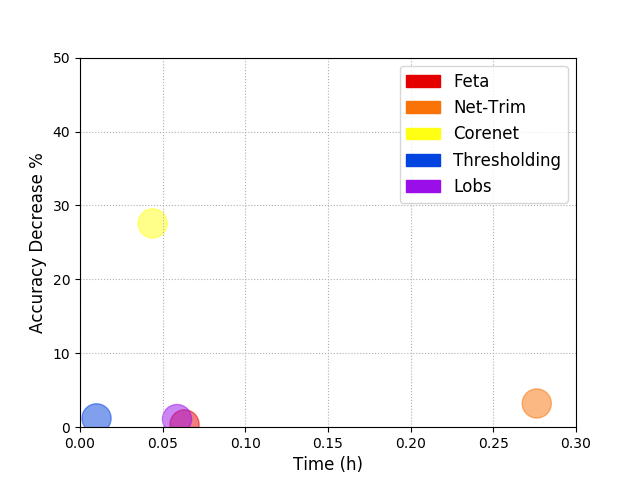}
        \caption{Conv Mnist}
    \end{subfigure}
    \begin{subfigure}[b]{0.3\linewidth}
        \includegraphics[scale=0.35]{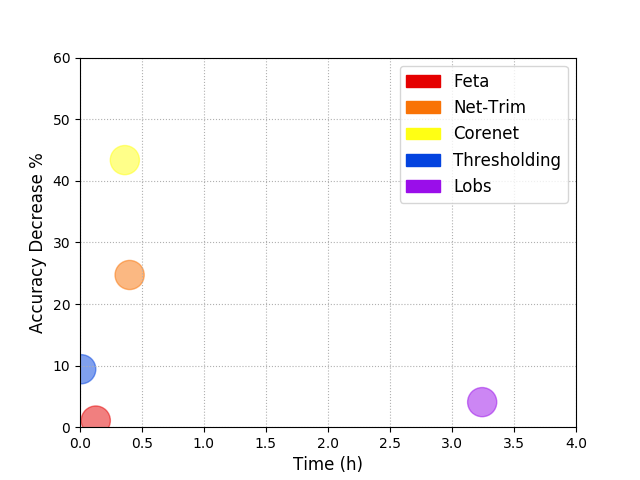}
        \caption{Conv Fashion Mnist}
    \end{subfigure}
    \begin{subfigure}[b]{0.3\linewidth}
        \includegraphics[scale=0.35]{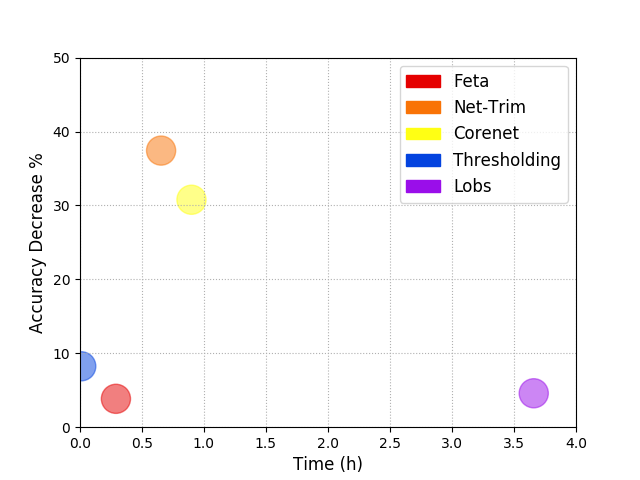}
        \caption{Conv Cifar}
    \end{subfigure}
    \caption{The test accuracy (\%) of pruned architectures vs the running time of the pruning algorithms in hours.  }
\end{figure*}

In order to solve the linearized problem we propose to use Accelerated Proximal SVRG (Acc-Prox-SVRG), which was presented in \cite{nitanda2014stochastic}. We detail this method in Algorithm 2. At each iteration a minibatch $\boldsymbol{A}'$ and $\boldsymbol{B}'$ is drawn. The gradient for the smooth part is calculated and the algorithm takes a step in that direction with step size $\eta$. Then the proximal operator for the non-smooth regulariser $\lambda \Omega(\cdot)$ is applied to the result. The hyperparameters for Acc-Prox-SVRG are the acceleration parameter $\beta$ and the gradient step $\eta$. We have found that in our experiments, using $\beta = 0.95$ and $\eta \in \{0.001 , 0.0001 \}$ gives the best results.

Our proposed algorithm has time complexity $\mathcal{O}(K(N+\frac{Nk}{N+\sqrt{k}}) \text{log}(\frac{1}{\epsilon}) d_1 d_2)$, where $\epsilon$ is the precision of the solution, $k$ is related to the Lipschitz and strong convexity constants, $d_2 \ll d_1$ and $K$ is the outer iteration number compared to $\mathcal{O}(md_1^3)$ for Net-trim. Furthermore the optimisation can be done in a stochastic manner using data minibatches greatly reducing the algorithm space complexity. We name our algorithm FeTa, Fast and Efficient Trimming Algorithm.

\section{Experiments without retraining.}

We now present detailed comparison experiments for pruning DNNs without retraining. The experimental setup (datasets and architectures) is identical to the one of Section 2. We prune both fully connected layers of all architectures to $90\%$ sparsity and plot the tradeoff between accuracy degradation and time complexity in Figure 2. The first thing that we notice is that Corenet and Net-Trim have inconsistent performance with regards to hard thresholding. While in some experiments their results are better or equal to hard thresholding in others they are significantly worse. On the other hand the FeTa and LOBS enjoy significant improvements in accuracy after pruning, while FeTa is up to $8\times$ faster than LOBS in some experiments. We also notice that the gains over hard thresholding are significant in fully connected architectures. On the contrary fully connected layers in convolutional DNN architectures are extremely redundant, with the majority of the non-linear separation of the data manifold being performed by the convolutional layers. Overall FeTa significantly outperforms Hard Thresholding and other smart pruning approaches both in final test accuracy and pruning speed, when retraining is not possible.

\section{Generalization Error}

We have already seen that hard thresholding followed by retraining is a very efficient pruning method. It is also interesting to explore theoretically what will be the response of the network after hard thresholding different layers at different sparsity levels. In the following section we use tools from the robustness framework \cite{xu2012robustness} to bound the generalization error of the new architecture induced by hard thresholding.

We need the following two definitions of the classification margin and the score that we take from \cite{sokolic2017robust}. These will be useful later for measuring the generalization error.

\begin{definition}
(\normalfont{Score}). For a classifier $g(\boldsymbol{x})$ a training sample $s_i = (\boldsymbol{x}_i,y_i)$ has a score

\begin{equation}
o(s_i)=o(\boldsymbol{x}_i,g(\boldsymbol{x}_i))=\min_{j \neq g(\boldsymbol{x}_i)}\sqrt{2}(\delta_{g(\boldsymbol{x}_i)}-\delta_{j})^{T}f(\boldsymbol{x}_i),
\end{equation}

where $\delta_i \in \mathcal{R}^{N_y}$ is the Kronecker delta vector with $(\delta_i)_i=1$, and $g(\boldsymbol{x}_i)$ is the output class for $s_i$ from classifier $g(\boldsymbol{x})$ which can also be $g(\boldsymbol{x}_i) \neq y_i$.
\end{definition}

\begin{definition}
(\normalfont{Training Sample Margin}). For a classifier $g(\boldsymbol{x})$ a training sample $s_i = (\boldsymbol{x}_i,y_i)$ has a classification margin $\gamma(s_i)$ measured by the $l_2$ norm if
\begin{equation}
g(\boldsymbol{x})=g(\boldsymbol{x}_i); \;\;\; \forall \boldsymbol{x} : ||\boldsymbol{x}-\boldsymbol{x}_i||_2< \gamma(s_i).
\end{equation}
\end{definition}

The classification margin of a training sample $s_i$ is the radius of the largest metric ball (induced by the $l_2$ norm) in $\mathcal{X}$ centered at $\boldsymbol{x}_i$ that is contained in the decision region associated with the classification label $g(\boldsymbol{x}_i)$. Note that it is possible for a classifier to misclassify a training point $g(\boldsymbol{x}_i) \neq y_i$. 



We are now ready to state our main result.

\begin{theorem}
Assume that $\mathcal{X}$ is a (subset of) $C_M$-regular k-dimensional manifold, where $\mathcal{N}(\mathcal{X};d;\rho) \leq (\frac{C_M}{\rho})^k$. Assume also that the DNN classifier $g_1(\boldsymbol{x})$ achieves a lower bound to the classification score $o(\tilde{s}) < o(s_i), \; \forall s_i \in S_m$ and take $l(g(\boldsymbol{x}_i),y_i)$ to be the $0-1$ loss. Furthermore assume that we prune classifier $g_1(\boldsymbol{x})$ on layer $i_{\star}$ using hard thresholding to obtain a new classifier $g_2(\boldsymbol{x})$. Then for any $\delta > 0$, with probability at least $1-\delta$, when $(\gamma-C_1 \cdot \frac{ \prod_{i > i_{\star}}||\boldsymbol{W}_i||_2}{ \prod_i||\boldsymbol{W}_i||_2}) > 0$,

\begin{equation}
\text{GE}(g_2) \leq A \cdot (\gamma-C_1\frac{ \prod_{i>i_{\star}}||\boldsymbol{W}_i||_2}{ \prod_{i}||\boldsymbol{W}_i||_2})^{-\frac{k}{2}}+B,
\end{equation}
where $A = \sqrt{ \frac{\log{(2)} \cdot N_y \cdot 2^{k+1} \cdot (C_M)^k}{ m } }$ and $B = \sqrt {\frac{2\log{1/\delta}}{m}}$ can be considered constants related to the data manifold and the training sample size, $\gamma = \frac{o(\tilde{s})}{\prod_i ||\boldsymbol{W}_i||_2 }$ is the margin and $C_1 = \max_{j\in\mathcal{S}_m}||f_{i_{\star}}(\boldsymbol{x}_j,\boldsymbol{U}_{i_{\star})}-f_{i_{\star}} (\boldsymbol{x}_j,\boldsymbol{W}_{i_{\star}})||_2$ is the maximum $l_2$ error of the layer over the training set.
\end{theorem}

The detailed proof can be found in Appendix A. The bound depends on two constants related to intrinsic properties of the data manifold, the regularity constant $C_M$ and the intrinsic data dimensionality $k$. In particular the bound depends exponentially on the intrinsic data dimensionality $k$. Thus more complex datasets are expected to lead DNNs that are less robust to pruning. The bound also depends on the spectral norm of the hidden layers $||\boldsymbol{W}_i ||_2$. Small spectral norms lead to a larger base in $(\cdot)^{-\frac{k}{2}} $ and thus to tigher bounds. 

Our result is quite pessimistic as the hard thresholding error $C_1$ is multiplied by the factor $\prod_{i > i_{\star}}||\boldsymbol{W}_i||_2$. Thus in our analysis the GE grows exponentially with respect to the remaining layer depth of the pertubated layer. This is in line with previous work \cite{raghu2016expressive} \cite{han2015learning} that demonstrates that layers closer to the input are much less robust compared to layers close to the output. 

We can use a perturbation bound introduced in \cite{neyshabur2017pac} to extend the above bound to include pruning of multiple layers 

\begin{theorem}
Assume that $\mathcal{X}$ is a (subset of) $C_M$-regular k-dimensional manifold, where $\mathcal{N}(\mathcal{X};d;\rho) \leq (\frac{C_M}{\rho})^k$. Assume also that the DNN classifier $g_1(\boldsymbol{x})$ achieves a lower bound to the classification score $o(\tilde{s}) < o(s_i), \; \forall s_i \in S_m$ and take $l(g(\boldsymbol{x}_i),y_i)$ to be the $0-1$ loss. Furthermore assume that we prune classifier $g_1(\boldsymbol{x})$ on all layers using hard thresholding, to obtain a new classifier $g_2(\boldsymbol{x})$. Then for any $\delta > 0$, with probability at least $1-\delta$, when $(\gamma-e D^2\sum_i\frac{||\boldsymbol{H}_i||_2}{ ||\boldsymbol{W}_i||_2}) > 0$,

\begin{equation}
\text{GE}(g_2) \leq A \cdot (\gamma-e D^2\sum_i\frac{||\boldsymbol{H}_i||_2}{ ||\boldsymbol{W}_i||_2})^{-\frac{k}{2}}+B,
\end{equation}
where $A = \sqrt{ \frac{\log{(2)} \cdot N_y \cdot 2^{k+1} \cdot (C_M)^k}{ m } }$ and $B = \sqrt {\frac{2\log{1/\delta}}{m}}$ can be considered constants related to the data manifold and the training sample size, $\gamma = \frac{o(\tilde{s})}{\prod_i ||\boldsymbol{W}_i||_2 }$ is the margin and $\boldsymbol{H}_i$ is the per layer perturbation matrix induced by hard thresholding.
\end{theorem}

\begin{figure*}[t!]
    \centering
    \begin{subfigure}[b]{0.3\linewidth}
        \includegraphics[scale=0.35]{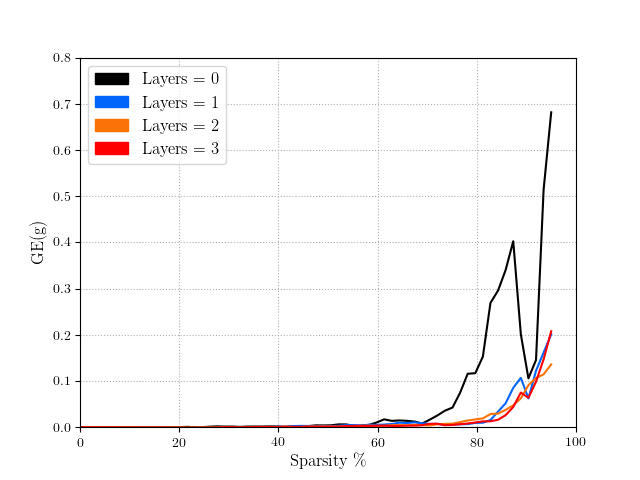}
        \caption{}
    \end{subfigure}
    \begin{subfigure}[b]{0.3\linewidth}
        \includegraphics[scale=0.35]{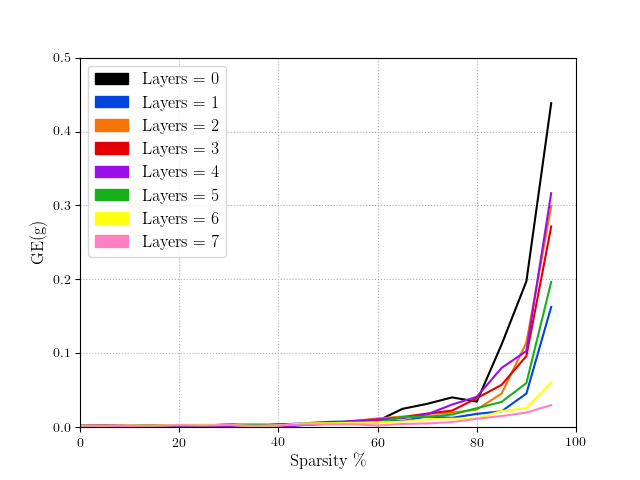}
        \caption{}
    \end{subfigure}
    \begin{subfigure}[b]{0.3\linewidth}
        \includegraphics[scale=0.35]{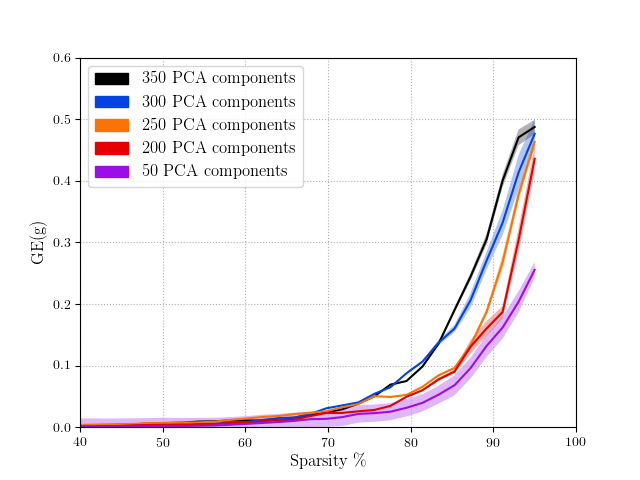}
        \caption{}
    \end{subfigure}
    \caption{a)Mnist convolutional b)Cifar convolutional c)Cifar convolutional for different PCA values. Layer "0" corresponds to the input layer.}
\end{figure*}

The detailed proof can be found in Appendix B. We note also the generality of our result; even though we have assumed a specific form of pruning, the GE bound holds for any type of bounded perturbation to a hidden layer. 

We make here an important note regarding generalization bounds for neural networks in general. All state of the art recent results such as \cite{neyshabur2017pac} \cite{bartlett2017spectrally} \cite{arora2018stronger} \cite{golowich2018size} provide only vacuous estimates of the generalization error. At the same time while the values of the generalization error estimated by these techinques are loose by orders of magnitude they still provide meanigfull insights into various properties of deep neural networks. In this context we will validate our theoretical framework by showing remaining layer depth correlates with an exponential decrease in DNN test accuracy. Also we will show that increased intrinsic data dimentionality correlates with decreased DNN robustness to pruning.  

Finally in the statement of our theorem we assume that $(\gamma- \{ \cdot \}) > 0$ this implies that all training samples remain correctly classified after pruning. What happens if this quantity goes to zero? In this case we can imagine the following three step procedure. We first start from a clean network and prune it and use our formula until the above quantity goes to zero. We can then recompute all relevant quantities for our bound including accuracy over the training set, margins, scores and spectral norms. We can then reapply our formula as the the quantity $(\gamma-\{ \cdot \})$ is by definition positive for a small enough pruning. Importantly the rate of margin decrease remains the same for all sparsity levels.

\section{Experiments on Generalization Error}

\subsection{Pruning different layers}
In this experiment we use the convolutional MNIST and the convolutional CIFAR architectures from section 2. We prune individual layers from the two networks using hard thresholding for different sparsity levels from $0\%$ to $95\%$ sparsity and compute the network accuracy.  We plot the results in Figure 3. We see that as predicted by our bound the network accuracy drops exponentially with remaining layer depth (the number of layers from the pruned layer to the output). This exponential behaviour is the result of the term $\prod_{i > i_{\star}}||\boldsymbol{W}_i||_2 = A^{|i>i_{\star}|} \;\; \text{if} \;\; ||\boldsymbol{W}_i||_2=A \;\; \forall i$, and not the result of the dimensionality exponent factor $k$. This suggests that if one has to chose it is beneficial to prune the layers close to the output first as this is have a minimal impact on the network accuracy. 

\subsection{Manifold dimensionality and pruning}
Our theoretical analysis suggests also that data with higher intrinsic dimensionality will lead to networks that are less robust to pruning. We test this hypothesis on the convolutional CIFAR architecture. We upper bound the intrinsic dimensionality of the data manifold by applying PCA for different number of latent dimensions, and then train different classification networks. Then we pruned using hard thresholding only the first convolutional layer and computed the accuracy over the test set. We average the results over 10 different networks for each upper bound. Denoting the dimensionality upper bound by $k$ we see that for $k=350$ the drop in accuracy is significantly larger compared to $k=50$. The degradation is not exactly exponential as our dimensionality factor suggests, however this is expected as our theoretical bound involves also other quantities such as the margin $\gamma$ and the spectral norm of different layers $||\boldsymbol{W}_i||_2$ which change for each trained network and will also affect the resulting accuracy.

\section{Conclusion}
We have seen that hard thresholding followed by retraining remains the most efficient way of pruning fully connected DNN layers. For the case without retraining we have introduced a novel algorithm called FeTa that is often orders of magnitude faster than competing approaches while maintaining network accuracy. We have also shown theoretically and empirically that it is more profitable to prune layers close to the output as the perturbation that is introduced results in a much smaller drop in accuracy compared to pruning earlier layers. We discover also that networks trained on data that lies on a manifold of high intrinsic dimensionality exhibit significantly reduced robustness to pruning.

\clearpage

\onecolumn

\section*{Appendix}

\subsection*{A. Proof of theorem 5.1.}
We will proceed as follows. We first introduce some prior results which hold for the general class of robust classifiers. We will then give specific prior generalization error results for the case of classifiers operating on datapoints from $C_m$-regular manifolds. Afterwards we will provide prior results for the specific case of DNN classifiers. Finally we will prove our novel generalization error bound and provide a link with prior bounds.  

\vspace{0.3 cm}

We first formalize robustness for generic classifiers $g(\boldsymbol{x})$. In the following we assume a loss function $l(g(\boldsymbol{x}),y)$ that is positive and bounded $0 \leq l(g(\boldsymbol{x}),y) \leq M$.
\begin{definition}
An algorithm $g(\boldsymbol{x})$ is $(K,\epsilon (\mathcal{S}_m))$ robust if $\mathcal{S}$ can be partitioned into K disjoint sets, denoted by $\{T_t\}_{t=1}^K$, such that $\forall s_i \in \mathcal{S}_m$, $\forall s \in \mathcal{S}$,

\begin{equation}
s_i,s \in T_t, \Rightarrow |l(g(\boldsymbol{x}_i),y_i)-l(g(\boldsymbol{x}),y)| \leq \epsilon(\mathcal{S}_m).
\end{equation}
\end{definition}

Now let $\hat{l}(\cdot)$ and $l_{\text{emp}}(\cdot)$ denote the expected error and the training error, i.e,

\begin{equation}
\hat{l}(g) \triangleq \mathbb{E}_{s \sim S}l(g(\boldsymbol{x}),y); \; \; \; l_{\text{emp}}(g) \triangleq \frac{1}{m} \sum_{s_i \in \mathcal{S}_m}l(q(\boldsymbol{x}_i),y_i)
\end{equation}

we can then state the following theorem from \cite{xu2012robustness}:

\begin{theorem}
If $\mathcal{S}_m$ consists of $m$ i.i.d. samples, and $g(\boldsymbol{x})$ is $(K,\epsilon (\mathcal{S}_m))$-robust, then for any $\delta > 0$, with probability at least $1-\delta$,

\begin{equation}
GE(g)=|\hat{l}(g) - l_{\text{emp}}(g)| \leq \epsilon (\mathcal{S}_m) + M \sqrt{\frac{2K\text{ln}2+2\text{ln}(1/ \delta )}{m} }.
\end{equation}

\end{theorem}

The above generic bound can be specified for the case of $C_m$-regular manifolds as in \cite{sokolic2017robust}. We recall the definition of the sample margin $\gamma(s_i)$ as well as the following theorem:

\begin{theorem}
If there exists $\gamma$ such that

\begin{equation}
\gamma(s_i) > \gamma > 0 \; \forall s_i \in S_m,
\end{equation}
then the classifier $g(\boldsymbol{x})$ is $(N_{\mathcal{Y}} \cdot \mathcal{N}(\mathcal{X};d,\gamma / 2 ),0 )$-robust.
\end{theorem}

By direct substitution of the above result and the definiton of a $C_m$-regular manifold into Theorem 0.1 we get: 

\begin{corollary}
Assume that $\mathcal{X}$ is a (subset of) $C_M$ regular $k-$dimensional manifold, where $\mathcal{N}(\mathcal{X};d,\rho)\leq(\frac{C_M}{\rho})^k$. Assume also that classifier $g(\boldsymbol{x})$ achieves a classification margin $\gamma$ and take $l(g(\boldsymbol{x}),y)$ to be the $0-1$ loss. Then for any $\delta > 0$, with probability at least $1-\delta$,
\begin{equation}
GE(g) \leq \sqrt{\frac{\text{log}(2)\cdot N_{\mathcal{Y}} \cdot 2^{k+1} \cdot (C_M)^k }{\gamma^k m} }+\sqrt{\frac{2 \text{log}(1/\delta) }{m} }.
\end{equation}

\end{corollary}

Note that in the above we have used the fact that $l(g(\boldsymbol{x}),y) \leq 1$ and therefore $M=1$. The above holds for a wide range of algorithms that includes as an example SVMs. We are now ready to specify the above bound for the case of DNNs, adapted from \cite{sokolic2017robust},

\begin{theorem}
Assume that a DNN classifier $g(\boldsymbol{x})$, as defined in equation 8, and let $\tilde{\boldsymbol{x}}$ be the training sample with the smallest score $o(\tilde{s})>0$. Then the classification margin is bounded as
\begin{equation}
\gamma(s_i)\geq \frac{o(\tilde{s})}{\prod_i ||\boldsymbol{W}_i||_2 } = \gamma.
\end{equation}
\end{theorem}

We now prove our main result. We want to relate the generalization error of an original unpruned classifier $g_1(\boldsymbol{x})$ to that of a new pruned classifier $g_2(\boldsymbol{x})$ at layer $i^{\star}$. Our analysis will be based on $C_1 = \max_{j\in\mathcal{S}_m}||f_{i_{\star}} (\boldsymbol{x}_j,\boldsymbol{W}_{i_{\star}})-f_{i_{\star}}(\boldsymbol{x}_j,\boldsymbol{U}_{i_{\star})}||_2 = ||f_{i_{\star}}^1(\boldsymbol{x}_j)-f_{i_{\star}}^2(\boldsymbol{x}_j)||_2$ the maximum $l_2$ norm error of the layer over the training set, after pruning. In the notation we have omitted the layer weights and introduced instead a numbered superscript $1$ and $2$ denoting unpruned and pruned layers respectively. We will denote by $ \tilde{\boldsymbol{x}} = \text{arg min}_{s_i \in S_m}\text{min}_{j \neq g(\boldsymbol{x}_i)} \boldsymbol{v}^{T}_{g(\boldsymbol{x}_i) j} f(\boldsymbol{x}_i) $ the training sample with the smallest score. For this training sample we will denote $j^{\star} = \text{arg min}_{j \neq g(\tilde{\boldsymbol{x}})} \boldsymbol{v}^{T}_{g(\tilde{\boldsymbol{x}}) j} f(\tilde{\boldsymbol{x}})$ the second best guess of the classifier $g(\cdot)$. Throughout the proof, we will use the notation $\boldsymbol{v}_{ij}=\sqrt{2}(\boldsymbol{\delta}_i-\boldsymbol{\delta}_j)$. 

\vspace{0.3 cm}

First we assume the score $o_1(\tilde{\boldsymbol{x}},g_1(\tilde{\boldsymbol{x}}))$ of the point $\tilde{\boldsymbol{x}}$ for the original classifier $g_1(\boldsymbol{x})$. Then, for the second classifier $g_2(\boldsymbol{x})$, we take a point $\boldsymbol{x}^{\star}$ that lies on the decision boundary between $g_2(\tilde{\boldsymbol{x}})$ and $j^{\star}$ such that $o_2(\boldsymbol{x}^{\star},g_2(\tilde{\boldsymbol{x}}))=0$. We assume for simplicity that, after pruning, the classification decisions do not change such that $g_1(\tilde{\boldsymbol{x}}) = g_2(\tilde{\boldsymbol{x}})$. We then make the following calculations

\begin{equation}
\begin{split}
o_1(\tilde{\boldsymbol{x}},g_1(\tilde{\boldsymbol{x}})) & = o_1(\tilde{\boldsymbol{x}},g_1(\tilde{\boldsymbol{x}})) - o_2(\boldsymbol{x}^{\star},g_2(\tilde{\boldsymbol{x}})) = \boldsymbol{v}^{T}_{g_1(\tilde{\boldsymbol{x}}) j^{\star}}f^1(\tilde{\boldsymbol{x}})-\boldsymbol{v}^{T}_{g_2(\tilde{\boldsymbol{x}}) j^{\star}}f^2(\boldsymbol{x}^{\star}) \\
& = \boldsymbol{v}^{T}_{g_2(\tilde{\boldsymbol{x}}) j^{\star}}(f^1(\tilde{\boldsymbol{x}})-f^2(\boldsymbol{x}^{\star})) \\
& \leq ||\boldsymbol{v}^{T}_{g_2(\tilde{\boldsymbol{x}}) j^{\star}}||_2||f^1(\tilde{\boldsymbol{x}})-f^2(\boldsymbol{x}^{\star})||_2 = ||f^1_L(\tilde{\boldsymbol{x}})-f^2_L(\boldsymbol{x}^{\star})||_2 \\
& \leq \prod_{i>i^{\star}} ||\boldsymbol{W}_i||_2||f^1_{i^{\star}}(\tilde{\boldsymbol{x}})-f^2_{i^{\star}}(\boldsymbol{x}^{\star})||_2 \\
& \leq \prod_{i>i^{\star}} ||\boldsymbol{W}_i||_2 \{ ||f^1_{i^{\star}}(\tilde{\boldsymbol{x}})-f^1_{i^{\star}}(\boldsymbol{x}^{\star})||_2 + ||f^1_{i^{\star}}(\boldsymbol{x}^{\star})-f^2_{i^{\star}}(\boldsymbol{x}^{\star})||_2 \} \\
& \leq \prod_{i>i^{\star}} ||\boldsymbol{W}_i||_2 \{ ||f^1_{i^{\star}}(\tilde{\boldsymbol{x}})-f^1_{i^{\star}}(\boldsymbol{x}^{\star})||_2 + C_1 \} \\
& \leq \prod_{i} ||\boldsymbol{W}_i||_2  ||\tilde{\boldsymbol{x}}-\boldsymbol{x}^{\star}||_2 + C_1\prod_{i>i^{\star}} ||\boldsymbol{W}_i||_2  \\
& \leq \prod_{i} ||\boldsymbol{W}_i||_2  \gamma_2(s_i) + C_1\prod_{i>i^{\star}} ||\boldsymbol{W}_i||_2, \\
\end{split}
\end{equation}

. From the above we can therefore write

\begin{equation}
\frac{o_1(\tilde{\boldsymbol{x}},g_1(\tilde{\boldsymbol{x}}))-C_1\prod_{i>i^{\star}} ||\boldsymbol{W}_i||_2}{\prod_{i} ||\boldsymbol{W}_i||_2} \leq  \gamma_2(\tilde{\boldsymbol{x}}).
\end{equation}

By following the derivation of the margin from the original paper \cite{sokolic2017robust} and taking into account the definition of the margin we know that

\begin{equation}
\gamma = \frac{o_1(\tilde{\boldsymbol{x}},g_1(\tilde{\boldsymbol{x}}))}{ \prod_{i} ||\boldsymbol{W}_i||_2} \leq \gamma_1(\tilde{\boldsymbol{x}}).
\end{equation}

Therefore we can finally write

\begin{equation}
\gamma - C_1\frac{\prod_{i>i^{\star}} ||\boldsymbol{W}_i||_2}{\prod_{i} ||\boldsymbol{W}_i||_2} \leq  \gamma_2(\tilde{\boldsymbol{x}}).
\end{equation}

The theorem follows from direct application of Corollary 0.2.1.

\subsection*{B. Proof of theorem 5.2.}
We start as in theorem 3.2 but instead of dealing with $C_1$ the maximum $l_2$ error over the training set we assume that pruning induces a perturbation matrix $\boldsymbol{H}_i$ to each layer $i$. We will then use the following stability result from \cite{neyshabur2017pac}.

\begin{theorem}
\cite{neyshabur2017pac}(Perturbation Bound). For any $D,L>0$, let $f_{\boldsymbol{w}} : \mathcal{X}_{D,N}\Rightarrow \mathbb{R}^k$ be a $L-$layer neural network with ReLU activations. Then for any $\boldsymbol{w}$, and $\boldsymbol{x}\in\mathcal{X}_{D,N}$, and any perturbation $\boldsymbol{h}=vec(\{\boldsymbol{H}_i\}_{i=0}^L)$ such that $||\boldsymbol{H}_i||_2 \leq \frac{1}{L}||\boldsymbol{W}_i||_2$, the change in the output of the network can be bounded as follows:

\begin{equation}
||f_{\boldsymbol{w}}(\boldsymbol{x})-f_{\boldsymbol{w}+\boldsymbol{h}}(\boldsymbol{x})||_2 = ||f^1(\boldsymbol{x})-f^2(\boldsymbol{x})||_2  \leq e D^2\prod_i ||\boldsymbol{W}_i||_2\sum_i\frac{||\boldsymbol{H}_i||_2}{ ||\boldsymbol{W}_i||_2}
\end{equation}

\end{theorem}

We can directly apply the above result to our case by assuming that layer pruning results in perturbation matrices $\boldsymbol{H}_i$ per layer $i$. 

We assume the score $o_1(\tilde{\boldsymbol{x}},g_1(\tilde{\boldsymbol{x}}))$ of the point $\tilde{\boldsymbol{x}}$ for the original classifier $g_1(\boldsymbol{x})$. Then, for the second classifier $g_2(\boldsymbol{x})$, we take a point $\boldsymbol{x}^{\star}$ that lies on the decision boundary between $g_2(\tilde{\boldsymbol{x}})$ and $j^{\star}$ such that $o_2(\boldsymbol{x}^{\star},g_2(\tilde{\boldsymbol{x}}))=0$. We assume as before that the classification decisions do not change such that $g_1(\tilde{\boldsymbol{x}}) = g_2(\tilde{\boldsymbol{x}})$. We write

\begin{equation}
\begin{split}
o_1(\tilde{\boldsymbol{x}},g_1(\tilde{\boldsymbol{x}})) & = o_1(\tilde{\boldsymbol{x}},g_1(\tilde{\boldsymbol{x}})) - o_2(\boldsymbol{x}^{\star},g_2(\tilde{\boldsymbol{x}})) = \boldsymbol{v}^{T}_{g_1(\tilde{\boldsymbol{x}}) j^{\star}}f^1(\tilde{\boldsymbol{x}})-\boldsymbol{v}^{T}_{g_2(\tilde{\boldsymbol{x}}) j^{\star}}f^2(\boldsymbol{x}^{\star}) \\
& = \boldsymbol{v}^{T}_{g_2(\tilde{\boldsymbol{x}}) j^{\star}}(f^1(\tilde{\boldsymbol{x}})-f^2(\boldsymbol{x}^{\star})) \\
& \leq ||\boldsymbol{v}^{T}_{g_2(\tilde{\boldsymbol{x}}) j^{\star}}||_2||f^1(\tilde{\boldsymbol{x}})-f^2(\boldsymbol{x}^{\star})||_2 = ||f^1(\tilde{\boldsymbol{x}})-f^2(\boldsymbol{x}^{\star})||_2 \\
& \leq ||f^1(\tilde{\boldsymbol{x}})-f^1(\boldsymbol{x}^{\star})||_2+||f^1(\boldsymbol{x}^{\star})-f^2(\boldsymbol{x}^{\star})||_2 \\
& \leq \prod_{i} ||\boldsymbol{W}_i||_2  ||\tilde{\boldsymbol{x}}-\boldsymbol{x}^{\star}||_2+e D^2\prod_i ||\boldsymbol{W}_i||_2\sum_i\frac{||\boldsymbol{H}_i||_2}{ ||\boldsymbol{W}_i||_2}\\
& \leq \prod_{i} ||\boldsymbol{W}_i||_2  \gamma_2(s_i)+e D^2\prod_i ||\boldsymbol{W}_i||_2\sum_i\frac{||\boldsymbol{H}_i||_2}{ ||\boldsymbol{W}_i||_2}
\end{split}
\end{equation}

We can then write

\begin{equation}
\frac{o_1(\tilde{\boldsymbol{x}},g_1(\tilde{\boldsymbol{x}}))- e D^2\prod_i ||\boldsymbol{W}_i||_2\sum_i\frac{||\boldsymbol{H}_i||_2}{ ||\boldsymbol{W}_i||_2}}{ \prod_i||\boldsymbol{W}_i||_2} \leq  \gamma_2(\tilde{\boldsymbol{x}}).
\end{equation}

Then as before

\begin{equation}
\gamma - e D^2\sum_{i}\frac{||\boldsymbol{H}_i||_2}{ ||\boldsymbol{W}_i||_2} \leq  \gamma_2(\tilde{\boldsymbol{x}}).
\end{equation}

The theorem follows from direct application of Corollary 0.2.1.

\subsection*{C. Experiment Details}

All architectures were trained using SGD with momentum $0.9$, learning rate of $0.01$, exponential decay and minibatch size of $32$. The architectures where trained until the validation accuracy saturated usually for $30$ epochs. 

We now give some details regarding the hyperparameters of the pruning algorithms. For all algorithms the entire training set was used to perform pruning. 

For Corenet algorithm we implemented neuron pruning (Corenet+) but did not implement amplification (Corenet++). Amplification increased linearly the computation time of the algorithm and we did not notice any improvements in accuracy for high sparsity levels. 

For Feta the algorithm was ran for $5$ to $12$ outer loups. The acceleration parameter was set to $\beta=0.95$, the SVRG learning rate was set to $\eta=0.001$ with minibatch size of $200$. In SVRG we need to set a number of full gradient computations denoted by the parameter $S$ we set this to $S=3$.

For Net-Trim we ran the algorithm for 400 iterations and $\rho=100$. We set the sparsity level by modifying the parameter $\epsilon$ as detailed in the authors instructions.

The LOBS argorithm does not contain any tunable parameters.

We did not implement GPU acceleration for any of the algorithms and it is not clear that this will provide speedups for common matrix-matrix and matrix-vector operations which are at the core of the implemented pruning algorithms. 

\bibliography{sample_paper}

\end{document}